\documentclass[10pt,twocolumn]{article}
\usepackage[a4paper,margin=18mm]{geometry}

\usepackage[english]{babel}
\usepackage[T1]{fontenc}
\usepackage[utf8]{inputenc}

\usepackage{graphicx}
\usepackage{bm}

\usepackage{amsmath,amssymb}

\usepackage{booktabs}
\usepackage{subcaption}
\usepackage{multirow}
\usepackage{adjustbox}

\usepackage{tikz}
\usetikzlibrary{arrows.meta,positioning,calc,backgrounds,fit,matrix,shapes}

\usepackage{etoolbox}
\usepackage{siunitx}

\usepackage[switch]{lineno} % switch = correct side per column

\usepackage{amsmath}
\pretocmd{\equation}{\linenomath}{}{}
\apptocmd{\endequation}{\endlinenomath}{}{}
\pretocmd{\align}{\linenomath}{}{}
\apptocmd{\endalign}{\endlinenomath}{}{}
\pretocmd{\gather}{\linenomath}{}{}
\apptocmd{\endgather}{\endlinenomath}{}{}

\title{Adaptive Sensing of Continuous Physical Systems for Machine Learning}

\author{
Felix K\"oster$^{1}$\thanks{Correspondence: felixk@mail.saitama-u.ac.jp}
\and
Atsushi Uchida$^{1}$\thanks{Correspondence: auchida@mail.saitama-u.ac.jp}
}

\date{
$^{1}$Department of Information and Computer Sciences, Saitama University,\\
255 Shimo-Okubo, Sakura-ku, Saitama City, Saitama, 338--8570, Japan\\[0.5em]
\today
}

\begin{document}
%\linenumbers
%\modulolinenumbers[1] % optional: label every 5th line
\maketitle

% ---------- Abstract ----------
\begin{abstract}
Physical dynamical systems can be viewed as natural information processors: their systems preserve, transform, and disperse input information. This perspective motivates learning not only from data generated by such systems, but also how to measure them in a way that extracts the most useful information for a given task. We propose a general computing framework for adaptive information extraction from dynamical systems, in which a trainable attention module learns both where to probe the system state and how to combine these measurements to optimize prediction performance. As a concrete instantiation, we implement this idea using a spatiotemporal field governed by a partial differential equation as the underlying dynamics, though the framework applies equally to any system whose state can be sampled. Our results show that adaptive spatial sensing significantly improves prediction accuracy on canonical chaotic benchmarks. This work provides a perspective on attention-enhanced reservoir computing as a special case of a broader paradigm: neural networks as trainable measurement devices for extracting information from physical dynamical systems.
\end{abstract}
\vspace{1em}
]

% ============================================================
\section{Introduction}
\label{sec:intro}

% Requires: \usepackage{tikz,adjustbox}
\begin{figure*}[t]
\centering
\begin{adjustbox}{width=\textwidth}
\begin{tikzpicture}[
  node distance=1.6cm,
  thick,
  every node/.style={transform shape},
  box/.style       ={rounded corners, rectangle, minimum width=2.2cm, minimum height=0.9cm, align=center},
  inputbox/.style  ={box, draw=blue!60,   fill=blue!10},
  pdebox/.style    ={box, draw=red!60,    fill=red!10,  minimum width=3.2cm},
  measbox/.style   ={box, draw=brown!70,  fill=brown!10, minimum width=3.8cm},
  attbox/.style    ={box, draw=purple!70, fill=purple!10, minimum width=5.0cm},
  wbox/.style      ={box, draw=purple!70, fill=purple!05, minimum width=3.8cm},
  outbox/.style    ={box, draw=green!60!black, fill=green!10, minimum width=2.6cm},
  trainbox/.style  ={box, draw=cyan!50!black, fill=cyan!10, minimum width=3.6cm},
  note/.style      ={font=\scriptsize, inner sep=1pt, fill=white}
]
\tikzset{>=latex}

% Main chain
\node (cin)   [inputbox]                                   {Input $\mathbf{x}_n$};
\node (pde)   [pdebox,  right=1.0cm of cin]                {Continuous reservoir\\ $u(\mathbf{x},t)$};
\node (meas)  [measbox, right=1.0cm of pde, text width=5cm]
              {Sampling \& projection\\[0.15em]
              $ \tilde{r}_t^{(j)}=\!\displaystyle\int \psi_t^{(j)}(\mathbf{x})\,u(\mathbf{x},t)\,d\mathbf{x}$\\
               $r_{t+T}^{(i)}=\!\displaystyle\int \phi_{t+T}^{(i)}(\mathbf{x})\,u(\mathbf{x},{t+T})\,d\mathbf{x}$};
\node (watt)  [wbox,   right=1.0cm of meas, text width=4.2cm]
              {Weighted sum\\[0.15em]
               $\mathbf{W}_{\mathrm{att},t+T}\mathbf{r}_{t+T}$};
\node (out)   [outbox, right=1.0cm of watt]                {Output $\bar{\mathbf{y}}_n$};

% Attention module above sampling
\node (att) [attbox, above=1.8cm of meas]
      {\textbf{Attention (trainable)}\\[0.2em]
       $F(\tilde{\mathbf r}_t)$ produces\\
       $\phi_{t+T}^{(i)}(\mathbf{x})$ (where to measure) and $\mathbf{W}_{\mathrm{att},t+T}$ (how to combine)};

% Gradient descent box above output
\node (gd) [trainbox, above=1.6cm of out]
      {Gradient descent\\ update of $F(\cdot)$};

% Forward connections
\draw[->] (cin)  -- (pde);
\draw[->] (pde)  -- (meas);

% meas -> weights -> output
\draw[->] (meas.east) -- ++(0.6,0) |- (watt.west);
\draw[->] (watt.east) -- (out.west);

% Attention -> weights (L-shaped)
\draw[->] (att.east) -| node[note, pos=0.25, yshift=6pt]{weights $\mathbf{W}_{\mathrm{att},t+T}$} (watt.north);

% Attention -> Sampling (downward kernels arrow, moved further right)
\draw[<-] (att.south west) ++(2.2,0) -- ++(0,-1.8)
    node[note, pos=0.55, xshift=3pt]{Input measurements $\tilde{\mathbf r}_t$};

% Sampling -> Attention (vertical return, slight right shift from kernels arrow)
\draw[<-] (meas.north east) ++(-0.2,0) -- ++(0,1.8)
    node[note, pos=0.45, xshift=3pt]{Output kernels $\phi_{t+T}^{(i)}(\mathbf{x})$};

% Output -> GD
\draw[->, dashed] (out.north) -- (gd.south);

% corners defined explicitly so it's UP, then LEFT, then DOWN
\coordinate (gdUp)    at ($(gd.north)+(0,23mm)$);      % go UP
\coordinate (aboveAtt) at ($(att.north)+(0,8mm)$);     % point ABOVE the top-center of att

\draw[->, dashed]
  (gd.north) -- (gdUp)     % UP
  -- (aboveAtt)            % LEFT/RIGHT horizontally (same y)
  -- (att.north);          % DOWN into top-center of att
\end{tikzpicture}
\end{adjustbox}

\caption{Flow chart of the Adaptive-Sensing-Attention–Enhanced-Reservoir-Computer (ASAERC). Inputs drive a fixed continuous reservoir. The reservoir is measured at fixed locations given via the measurement kernels $\psi_t^{(j)}(\mathbf{x})$ and the resulting $N_\mathrm{fix}$ states $\tilde{r}_t$ are fed into the attention module. The trainable attention module outputs parameterized kernels $\phi_{t+T}^{(i)}(\mathbf{x})$ to guide sampling and attention weights $\mathbf{W}_{\mathrm{att},t+T}$ to combine sampled features. The weighted combination produces $\bar{\mathbf{y}}_n$, and gradient descent (dashed path) updates the attention module. No gradients pass through the PDE reservoir.}
\label{fig:caerc-flow}
\end{figure*}
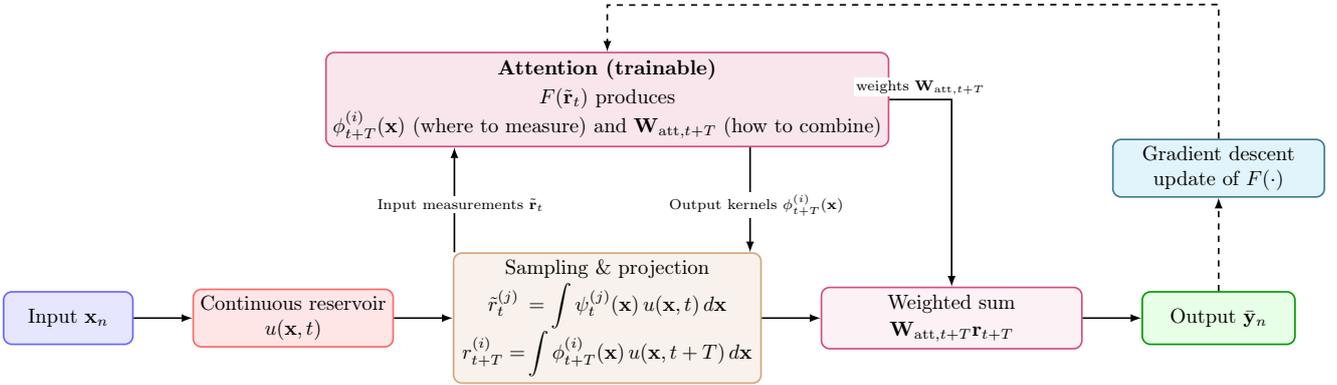

\begin{figure*}[t]
\centering
\includegraphics[width=0.95\textwidth]{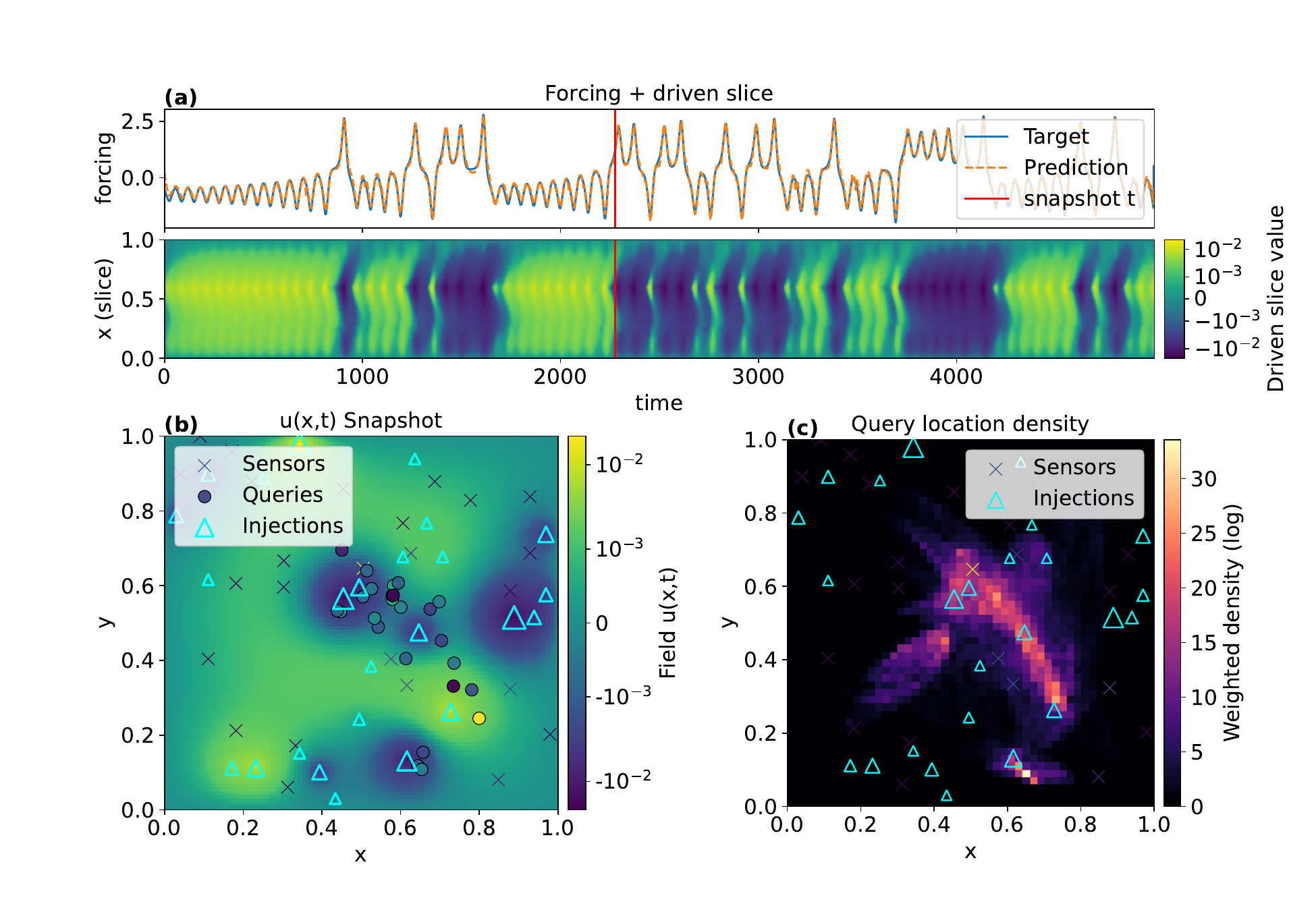}\\[1ex]
\caption{\textbf{Illustration of ASAERC behavior.}
(a) Example predicted time series (orange-dashed line) compared with ground truth (blue-solid line) on the Lorenz attractor. 
(b) Snapshot of the PDE reservoir field $u(\mathbf{x},t^\star)$ with fixed injection points (light-blue triangles), fixed measurement points (attention input, crosses) $\tilde{r}_t^{(j)} = \int_{\Omega} \psi_t^{(j)}(\mathbf{x})\, u(\mathbf{x},t)\, d\mathbf{x},$ and dynamic measurement locations (attention output) overlaid $r_{t+T}^{(j)} = \int_{\Omega} \phi_{t+T}^{(i)}(\mathbf{x})\, u(\mathbf{x},t+T)\, d\mathbf{x},$.
The adaptive sensors (blue circles) cluster around dynamically active regions, demonstrating the network’s ability to focus its attention on informative spatial locations. (c) A histogram of the attention output locations, giving an intricate pattern that is a nonlinear projection of the Lorenz attractor on the measurement process.}
\label{fig:caerc-behavior}
\end{figure*}

Dynamical systems naturally process information: their trajectories preserve, transform, and disperse information about external inputs. This idea was formalized through the concept of information processing capacity (IPC), which quantifies how well a dynamical system can reconstruct functions of its past inputs when coupled to a fixed output layer \cite{Dambre2012}. Treating dynamics as information reservoirs has motivated computational paradigms that exploit state evolution directly for prediction and control.

Reservoir computing (RC) is a prominent example. It uses a high-dimensional recurrent dynamical system as a fixed nonlinear feature generator and trains only a simple readout layer \cite{jaeger2001echo,appeltant2011information,tanaka2019recent,torrejon2017neuromorphic,kotooka2021}. RC can be computationally efficient, because only the readout is optimized and is well suited to time-series prediction, pattern recognition, and classification \cite{jaeger2001echo,tanaka2019recent}, while also being attractive for hardware implementation. RC has been demonstrated in a wide variety of physical platforms, including electronic delay systems \cite{appeltant2011information,Koester2021MMF}, photonic reservoirs \cite{appeltant2011information,Picco2025DelayedInputOpt}, spintronic devices \cite{torrejon2017neuromorphic}, soft robotic bodies \cite{ef0a6b769141456283377bec586ba791}, and nanostructure materials \cite{kotooka2021}. These examples emphasize that many complex media can serve as information-processing systems.

Nevertheless, classical RC has a key limitation: its readout is static. Reservoir states are projected to outputs via a fixed linear mapping, which can be too restrictive to exploit nonlinear information in the state space, especially in multiscale or chaotic regimes. Several approaches attempt to mitigate this rigidity, for instance by optimizing readout hyperparameters \cite{Flynn2021,Flynn2024} or injecting control signals to steer predictions \cite{Du2025}. However, such strategies often require task-specific tuning, rely on teacher forcing, or do not transfer robustly across diverse dynamical regimes.

In parallel, machine learning has introduced mechanisms for adaptive feature selection. Attention dynamically re-weights a representation based on its current context \cite{vaswani2017attention,bai2018empirical}. Building on this idea, attention-enhanced reservoir computing (AERC) replaces the static linear readout with a small neural network that computes state-dependent weights at each time step \cite{koester2024}. This enables a single trained model to represent multiple attractors and switch between them without retraining, which is crucial for autonomous multifunctionality.

In this work we propose a broader viewpoint: instead of adapting only the readout weights, we learn the measurement process itself. We introduce a framework for trainable information extraction from dynamical systems, in which a neural interface learns both \emph{where to measure} a system’s state and \emph{how to combine} the resulting measurements for prediction.
We propose a concept of Adaptive-Sensing Attention-Enhanced Reservoir Computing (ASAERC), which relates to classical sensor placement and experimental design, where one selects a small set of sensors to reconstruct or monitor a spatial state.
For example, data-driven sparse sensor placement methods choose fixed measurement locations that enable accurate state reconstruction in a learned low-dimensional basis \cite{Manohar2017}, and information-theoretic approaches select fixed sensor sets by maximizing mutual information under a Gaussian-process model \cite{Krause2008}.
In contrast to these predominantly offline and static designs, ASAERC learns a state-conditioned, time-varying measurement operator: the sensing kernels $\{\phi^{(i)}_{t+T}\}$ are produced online from the current fixed measurements $\tilde{\mathbf r}_t$ to directly minimize a forecasting loss.

Our approach is also related to learned sampling modules in deep learning, such as spatial transformer networks \cite{Jaderberg2015} and deformable convolutions \cite{Dai2017}, which adaptively resample feature maps to focus computation on task-relevant regions.
The key distinction is that ASAERC treats sensing as an interface to an external continuous dynamical system: we learn where to probe and how to combine measurements of a fixed reservoir field $u(\mathbf x,t)$, while explicitly avoiding backpropagation through the reservoir dynamics.
This reformulates attention-enhanced reservoir computing as a special case of a broader paradigm in which neural networks act as trainable measurement devices for continuous physical systems.

We demonstrate ASAERC on one-step-ahead prediction across a heterogeneous suite of chaotic systems spanning continuous-time flows, discrete maps, and a delay differential equation.
Across all systems, adaptive sensing improves prediction accuracy compared to classical RC and fixed-sensor attention baselines, supporting the view that learning \emph{where to measure} can be as important as learning \emph{how to combine} measurements.

\begin{figure}[t]
\centering
\includegraphics[width=0.5\textwidth]{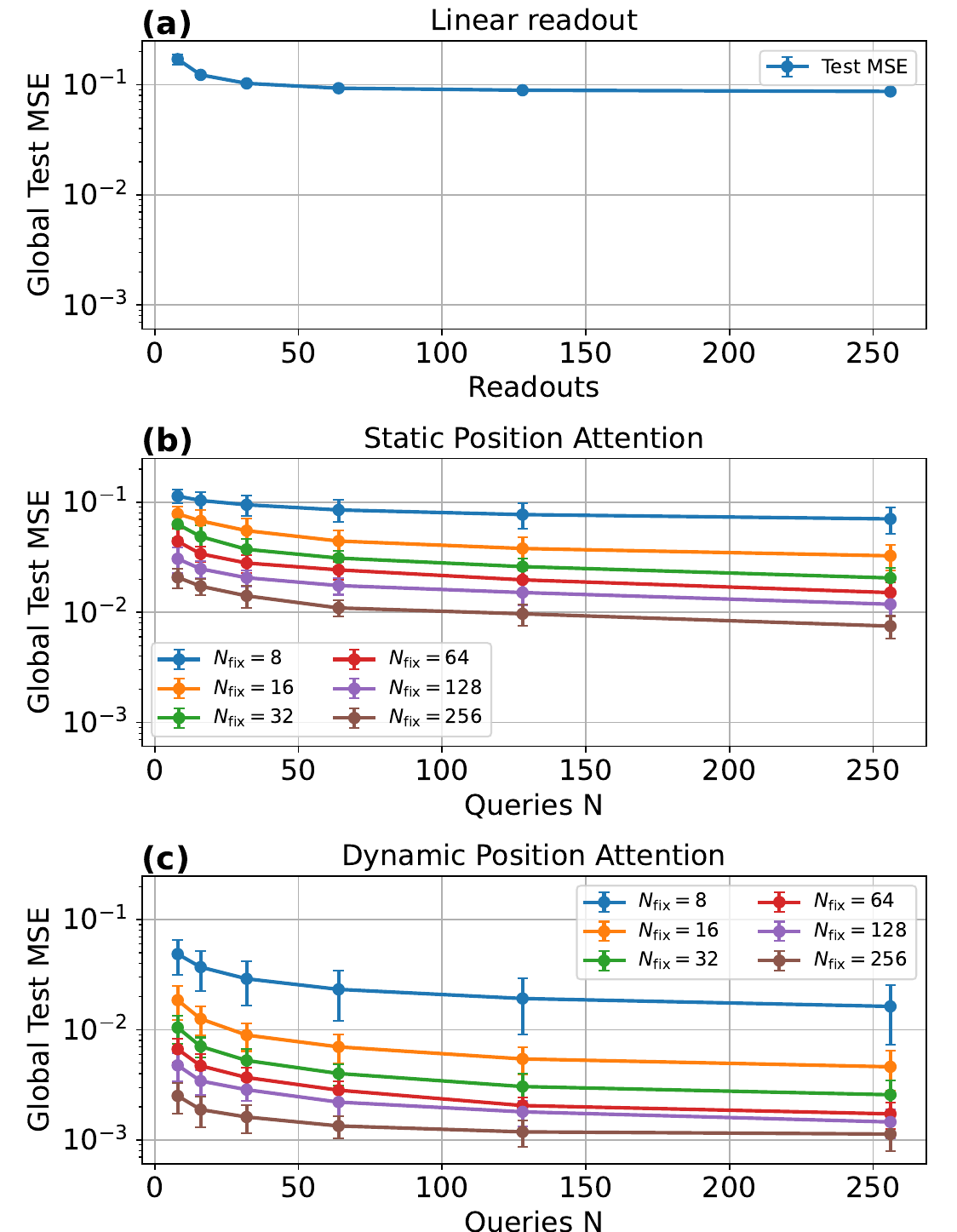}\hfill
\caption{\textbf{Prediction error as a function of number of measurement points.}
All methods use the same PDE reservoir.  
(a) Classical reservoir computing with linear readout; (b) AERC with fixed measurement positions; (c) ASAERC with adaptive sensing. Color indicates the number of measurement points at fixd locations.}
\label{fig:mse-vs-points}
\end{figure}

% ============================================================
\section{Results}
\label{sec:results}

\subsection{Scheme overview}
\label{subsec:overview}

Figure~\ref{fig:caerc-flow} summarizes the ASAERC pipeline.
A standardized discrete-time input $\mathbf{x}_n$ drives a fixed continuous reservoir field $u(\mathbf{x},t)$.
At each time step, fixed measurements $\tilde{\mathbf r}_t$ are collected via kernels $\{\psi_t^{(j)}\}$ and fed to a trainable attention module.
Conditioned on $\tilde{\mathbf r}_t$, the module outputs are the following: (i) adaptive measurement kernels $\{\phi^{(i)}_{t+T}\}$ that specify where to probe the field at a later time and (ii) attention weights $\mathbf{W}_{\mathrm{att},t+T}$ that combine the resulting measurements into the discrete predictions $\bar{\mathbf y}_n$.
Only the attention module is optimized by gradient descent; the reservoir dynamics are not differentiated (dashed arrows in Fig.~\ref{fig:caerc-flow}).

As a concrete instantiation of ASAERC, we use a two-dimensional diffusion-type field
$u(\mathbf{x},t)$ on a rectangular domain $\Omega=[0,L_x]\times[0,L_y]$,
driven by the input signal $f(I)$ through spatially localized sources: A model is described by a partial differential equation (PDE) as follows
\begin{align}
  \partial_t u(\mathbf{x},t) = \nu \Delta u(\mathbf{x},t) + f(I),
  \label{eq:pde-core-short}
\end{align}
with homogeneous Dirichlet boundary conditions $u=0$ on $\partial\Omega$.
The field is discretized on a uniform grid and integrated forward in time with a stable explicit scheme (see Methods).

At each step, we obtain two sets of scalar observations from the field.
First, a fixed set of $N_{\mathrm{fix}}$ measurements
$\tilde{\mathbf r}_t \in \mathbb{R}^{N_{\mathrm{fix}}}$
is collected at static locations (used as input to the attention module).
Conditioned on $\tilde{\mathbf r}_t$, the network outputs consists of: (i) $N$ adaptive measurement locations and
(ii) attention weights, producing an adaptive state $\mathbf r_{t+T}\in\mathbb{R}^{N}$
that is linearly combined to yield the prediction
$\bar{\mathbf y}_t=\mathbf W_{\mathrm{att},t+T}\mathbf r_{t+T}$.
Field values at off-grid locations are obtained by bilinear interpolation (see Methods).

We evaluate ASAERC on the task of one-step-ahead prediction across a heterogeneous dataset of eight canonical nonlinear systems:
five continuous-time systems (Lorenz system~\cite{lorenz1963},
R\"ossler system~\cite{rossler1976},
Van der Pol oscillator~\cite{vanderpol1926},
Duffing oscillator~\cite{duffing1918},
and the double pendulum~\cite{levien1993doublependulum}),
two discrete maps (logistic map~\cite{may1976},
H\'enon map~\cite{henon1976}),
and one delay differential equation (Mackey--Glass system~\cite{mackeyglass1977}).
This extends our earlier discrete AERC study~\cite{Koester2025} to a broader and more diverse benchmark suite.
Continuous-time systems are integrated numerically, and all trajectories are resampled to a common length for training and evaluation.
Sampling parameters for all systems are summarized in the Method secion.

\subsection{Qualitative behavior of ASAERC}
\label{subsec:qualitative}

We first illustrate how ASAERC performs adaptive sensing in a concrete example.
As summarized in Fig.~\ref{fig:caerc-flow}, the normalized input $\mathbf{x}_n$ drives a continuous reservoir field $u(\mathbf{x},t)$, which is measured at $N_\mathrm{fix}$ fixed locations via kernels $\psi_t^{(j)}(\mathbf{x})$. A trainable attention module $F(\mathbf{r}_t)$ receives the $N_\mathrm{fix}$ measured states $\tilde{r}_t^{(j)}$. After processing these inputs the module selects $N$ measurement locations $\phi_{t+T}^{(i)}(\mathbf{x})$, which we call queries, resulting in a new set of $N$ measured states $r_{t+T}^{(j)}$, and combines these samples through state-dependent weights $\mathbf{W}_{\mathrm{att},t+T}$. Here $T$ takes into account the time it needs for the processing of the information in the attention module and measurement procedure done via the $\phi_{t+T}^{(i)}(\mathbf{x})$ measurement device. The attention module is trained via a gradient descent algorithm which does not need any information about the reservoir.

Figure~\ref{fig:caerc-behavior} visualizes this process for a representative simulation driven by the Lorenz system. The setup for the reservoir can be found in Sec. \ref{sec:pde-example} of the Methods section.
Figure~\ref{fig:caerc-behavior}(a) shows a short segment of open-loop one-step-ahead prediction for one Lorenz coordinate.
The predicted trajectory $\hat{x}(n)$ (orange-dashed) closely follows the ground truth $x(n)$ (blue-solid), indicating that the sensed reservoir state contains sufficient information for accurate short-term forecasting.
The lower panel plots a diagonal one-dimensional slice through the two-dimensional reservoir field $u(\mathbf{x},t)$ over the same time window, providing a direct view of how the Lorenz forcing produces structured spatiotemporal activity in the PDE substrate.

Figure~\ref{fig:caerc-behavior}(b) visualizes the reservoir state at a representative time $t^\star$.
The heat map shows $u(\mathbf{x},t^\star)$ on the spatial domain.
Light-blue triangles mark the fixed injection sites through which the input signal $\mathbf{x}$ is written into the field (marker size is proportional to injection strength).
Crosses denote the fixed measurement locations $\{\psi_t^{(j)}(\mathbf{x})\}_{j=1}^{N_\mathrm{fix}}$ that provide the attention module with its input features $\tilde{\mathbf{r}}_t \in \mathbb{R}^{N_\mathrm{fix}}$ (cf.\ Fig.~\ref{fig:caerc-flow}).
Superimposed blue circles indicate the adaptive measurement locations chosen by the attention module at $T$ time-step after $\{\psi_t^{(j)}(\mathbf{x})\}_{j=1}^{N_\mathrm{fix}}$ were measured, i.e.\ the queries $\{\phi_{t+T}^{(i)}(\mathbf{x})\}_{i=1}^{N}$ that define the values collected into $\mathbf{r}_{t+T} \in \mathbb{R}^{N}$.
The color intensity reflects how strongly each adaptive measurement contributes to the final prediction through $\mathbf{W}_{\mathrm{att},t}$.
For clarity, we show a small configuration with 25 fixed sensors and 25 adaptive queries, chosen to make the sensing behaviour visually interpretable.

Figure~\ref{fig:caerc-behavior}(c) aggregates the adaptive sensing decisions over the full Lorenz trajectory by plotting a histogram of weighted query locations.
Rather than sampling uniformly, ASAERC repeatedly selects structured regions of the domain, revealing a learned measurement strategy that is shaped by both the reservoir dynamics and the target prediction task.
The queries tend to cluster around large injections and avoid the boundaries due to the chosen homogeneous Dirichlet boundary conditions $u=0$ on $\partial\Omega$.
Together, these results provide an intuitive interpretation of ASAERC as a trainable sensing interface: the model writes the input into a continuous system, adaptively probes the resulting field, and combines complementary measurements to produce accurate forecasts.

\subsection{Performance versus measurement points}
\label{subsec:performance-measurement-points}

We next quantify how predictive performance depends on the number of measurement points. We use eight different chaotic systems as data set. A detailed description can be found in Sec. \ref{subsec:benchmarks} of the Method section.
Figure~\ref{fig:mse-vs-points} reports the mean-squared error (MSE) over the full data set as a function of the number of output measurement points $\{\phi_t^{(i)}(\mathbf{x})\}_{i=1}^{N}$ for three methods:
(a) classical reservoir computing with linear readout,
(b) AERC with fixed measurement positions and adaptive weights, and
(c) ASAERC with adaptive sensing and adaptive weights.
All methods share the same PDE reservoir, so differences arise purely from the measurement and readout strategy.

The linear readout baseline in Fig.~\ref{fig:mse-vs-points}(a) uses a fixed linear mapping for prediction.
In this case, the input measurements $\{\psi_t^{(j)}(\mathbf{x})\}_{j=1}^{N_\text{fix}}$ and output measurements $\{\phi_{t+T}^{(i)}(\mathbf{x})\}_{i=1}^{N}$ are identical as no attention module is used.
The error remains comparatively large and saturates around $10^{-1}$, as expected since a single static readout must balance performance across multiple attractors.

The AERC approach in Figure~\ref{fig:mse-vs-points}(b) computes the readout weights dynamically with a small neural network, however it keeps the measurement positions fixed in space.
Only the weights adapt over time.
On the $x$-axis we vary the number of output queries $\{\phi_{t+T}^{(i)}(\mathbf{x})\}_{i=1}^{N}$, while different colors correspond to different numbers of fixed input measurements $\{\psi_t^{(j)}(\mathbf{x})\}_{j=1}^{N_\text{fix}}$.
The hidden layer size is fixed at 128 nodes.
This adaptive weighting already yields a strong improvement, reducing the error below $10^{-2}$ for sufficiently many fixed inputs.

The ASAERC approach in Fig.~\ref{fig:mse-vs-points}(c) makes the measurement positions adaptive.
Here, the attention module outputs both weights and measurement locations at each time step.
All hyperparameters are kept identical to the AERC configuration for a fair comparison.
ASAERC achieves an additional order-of-magnitude error reduction beyond AERC, confirming that optimizing where the system is sensed provides substantial gains beyond optimizing how the sensed values are combined.
With 256 measurement points, the overall error decreases to around $10^{-3}$, and even the 16-input ASAERC outperforms the 256-input AERC settings.

\subsection{Trainable parameter count versus number of measurement points}
\label{subsec:parameter-count}

\begin{figure}[t]
\centering
\includegraphics[width=0.5\textwidth]{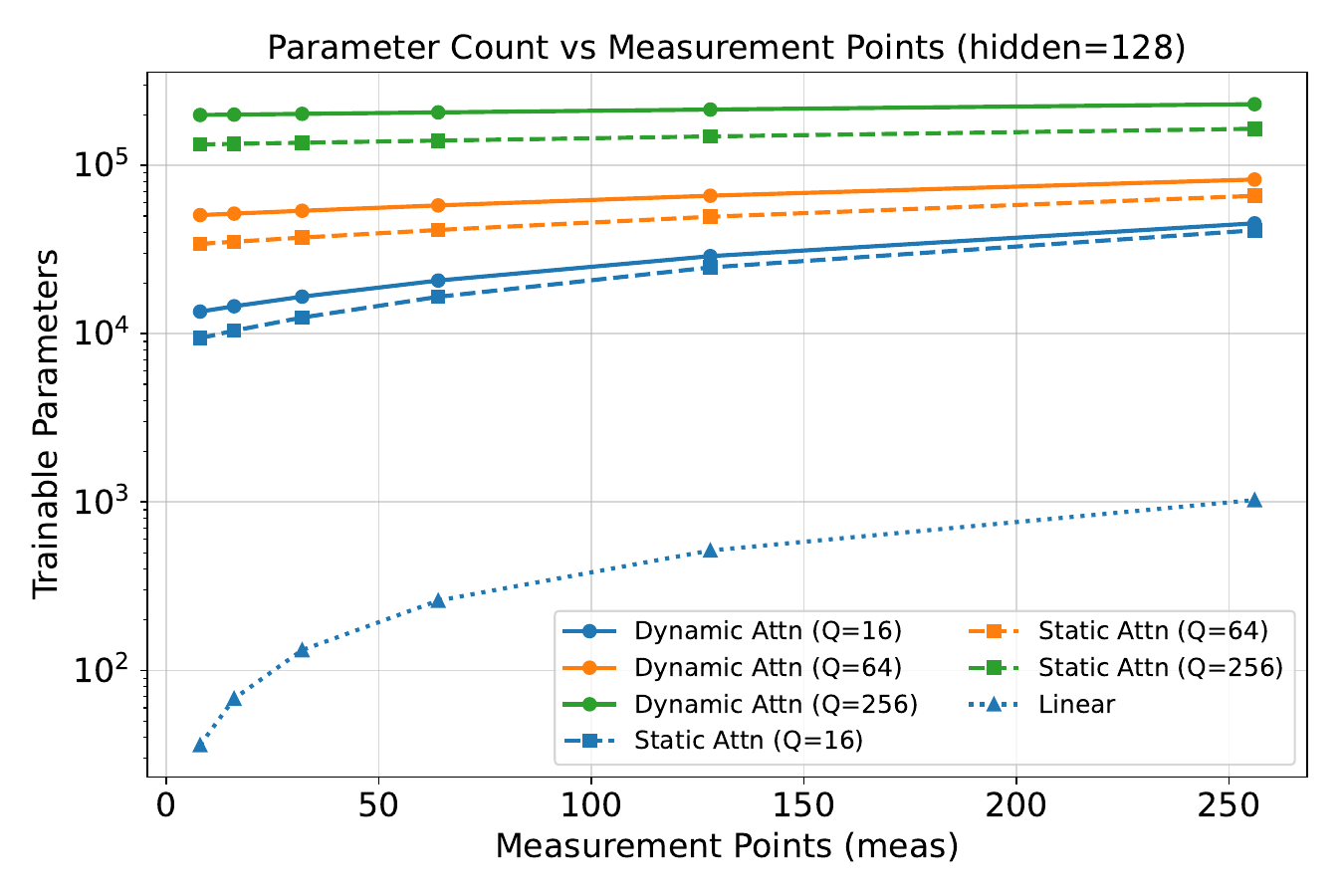}\hfill
\caption{\textbf{Number of trainable parameters versus number of measurement points.}
The dotted line with triangles denotes the linear readout, the dashed line with squares denotes AERC, and the solid lines with dots denote ASAERC.
For both AERC and ASAERC, we show three cases: one with 16, 64, and 256 adaptive queries. All three methods use the same PDE reservoir. ASAERC has the largest number of parameters, however only slightly more than AERC. Linear readout has far fewer parameters.}
\label{fig:param-vs-points}
\end{figure}

To assess model complexity, we compare the number of trainable parameters as a function of the number of fixed input measurements $\{\psi_t^{(j)}(\mathbf{x})\}_{j=1}^{N_\text{fix}}$.
Figure~\ref{fig:param-vs-points} summarizes parameter counts for linear readout (dotted line), AERC (dashed line), and ASAERC (solid line).

Linear readout scales linearly with the number of measurement points because only the output matrix is trained.
AERC requires more parameters due to the neural network mapping from fixed measurements to attention weights, and its parameter count grows with $N_\text{fix}$.
ASAERC adds a second output head to generate measurement locations, resulting in slightly more parameters than AERC.
However, this difference is essentially a constant offset: it corresponds to the additional linear layer required to predict kernel parameters.

Overall, AERC and ASAERC operate in $10^{4}$ to $10^{5}$ parameter regime, making training computationally trivial on modern hardware.
More importantly, ASAERC maintains substantially lower prediction errors than AERC even where their parameter counts are nearly identical (cf.\ Fig.~\ref{fig:mse-vs-points}).
This indicates that ASAERC’s gains stem from adaptive measurement placement rather than simply increasing model capacity.

\subsection{Correlation analysis of reservoir readouts}
\label{subsec:correlation-analysis}

\begin{figure*}[t]
\centering
\includegraphics[width=0.9\textwidth]{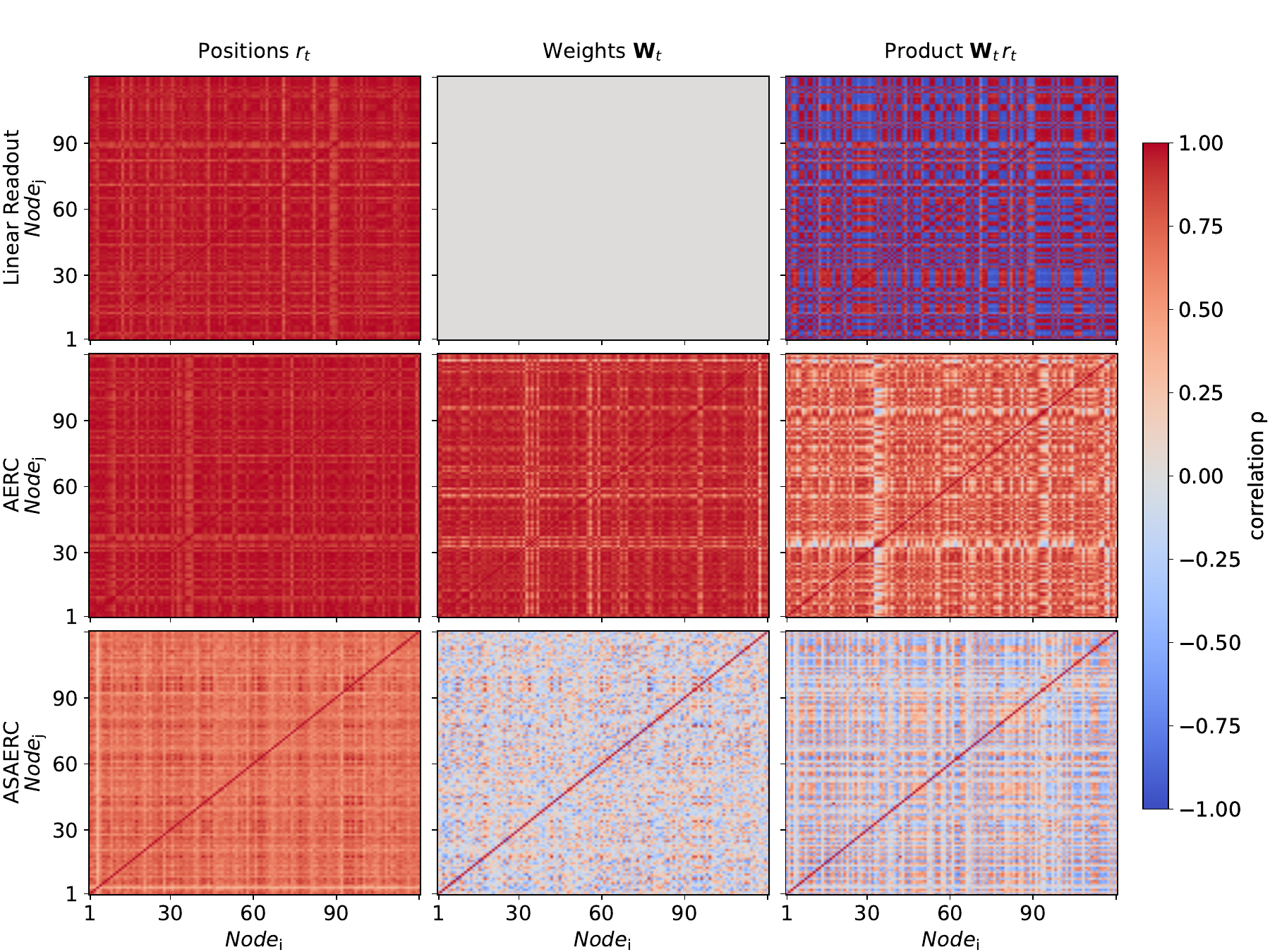}
\caption{\textbf{Correlation analysis of readout contributions.}
Distribution of pairwise Pearson correlation coefficients $\rho_{ij}$ between readout nodes $i$ and $j$ for (left column) raw PDE values, (middle) attention weights, and (right) their product.
Rows correspond to (top row) linear readout, (middle row) AERC, and (bottom row) ASAERC
ASAERC exhibits substantially lower correlation than linear readout and AERC, indicating that
its measurements are less redundant and more complementary, which likely aids generalization
and lowers prediction error.}
\label{fig:readout-correlation}
\end{figure*}

To further probe why ASAERC achieves improved predictive performance, we examine statistical dependence between readout nodes.
For each pair of measurement points $(i,j)$, we compute the Pearson correlation coefficient 
\begin{align}
  \rho_{ij} = \frac{\mathrm{Cov}(z_i,z_j)}{\sigma(z_i)\,\sigma(z_j)},
\end{align}
where $\mathrm{Cov}(z_i,z_j)$ is the covariance between the two time series and $\sigma(z_i)$ denotes the standard deviation of $z_i$ (computed over time and then averaged over trials as described below), where $z_i$ denotes the quantity associated with index $i$.

Figure~\ref{fig:readout-correlation} summarizes the distribution of correlation coefficients across all measurement pairs, averaged over the dataset.
We consider three cases:
(left column) raw PDE field values at measurement locations,
(middle) learned attention weights,
and (right) their product, i.e.\ the effective readout contribution. Rows correspond to (top row) linear readout, (middle row) AERC, and (bottom row) ASAERC.

In the linear readout baseline (top row), the raw sampled values exhibit strong pairwise correlations clustered near $\pm 1$, indicating substantial redundancy induced by fixed measurement locations. Their Pearson correlation across time is formally undefined, because the readout weights are time-invariant. For clarity, we set the corresponding weight--weight correlations to zero in Fig.~\ref{fig:readout-correlation}. The correlations of the effective contributions (weight $\times$ value) largely mirror those of the raw values, with an additional sign flip determined by the product of the corresponding readout weights.
In AERC (middle row), the learned weights reduce redundancy in the weight distribution, however correlations among nodes remain pronounced because measurement locations stay fixed.
In ASAERC (bottom row), correlations are substantially reduced, especially in the combined contribution.
A surprising result is that the adaptive weights reduce correlation significantly because positions can now also be learned.
This indicates that adaptive positioning selects less redundant measurements and promotes feature diversity.
The attention mechanism then allocates complementary weights, yielding a more balanced and informative representation.

From these results, correlation analysis suggests that ASAERC’s advantage is not merely due to flexible weights.
Its key contribution is adaptive reallocation of measurement positions, which reduces redundancy and enhances feature diversity in the sensed representation.

% ============================================================
\section{Conclusion}

In this work, we introduced Adaptive-Sensing Attention–Enhanced Reservoir Computer (ASAERC), a general framework for adaptive information extraction from continuous dynamical systems. ASAERC simultaneously learns \emph{where to measure} the underlying spatiotemporal field and \emph{how to combine} these measurements for prediction by combining a PDE-based reservoir with a trainable attention mechanism. This dual adaptivity reformulates reservoir computing from a paradigm of fixed feature extraction to one of trainable measurement, effectively turning neural networks into dynamic sensors of physical systems.

We demonstrated that ASAERC substantially outperforms both classical reservoir computing with linear readout and Attention-Enhanced Reservoir Computing (AERC) through systematic evaluation on eight benchmark systems. Adaptive sensing yielded prediction errors up to one order of magnitude lower than AERC and two orders lower than linear readouts, with only small increases in parameter count.

Correlation analysis further revealed that ASAERC reduces statistical dependence among readout nodes, leading to more diverse and complementary feature representations. Overall, these results highlight adaptive measurement as a powerful principle for extracting information from complex dynamical systems. While we instantiated ASAERC with a diffusion-type PDE reservoir, the framework is general and can be applied to other spatiotemporal physical systems for machine learning.

% ============================================================
\section{Methods}
\label{sec:methods}

\subsection{Classic Reservoir Computing}
\label{subsec:classic-rc}

Reservoir computing (RC) is a computational framework that employs a fixed, high--dimensional dynamical system, called a reservoir, in conjunction with a trainable readout layer to perform tasks such as time--series prediction, pattern recognition, and classification \cite{jaeger2001echo,appeltant2011information,tanaka2019recent}. Many different types of reservoirs can be used. The most common approach consists of randomly connected nonlinear nodes that transform the input into a rich dynamical representation, while only the linear output weights are optimized.

As an example, we can take the echo state network \cite{jaeger2001echo}. The reservoir state at time step \( t \), denoted by \( \mathbf{r}_t \), evolves according to
\begin{align}
   \mathbf{r}_t = \tanh\!\left(\mathbf{W}_{\text{res}} \mathbf{r}_{t-1} + \mathbf{W}_{\text{in}} \mathbf{x}_t + \mathbf{b}\right),
   \label{eq:rc-update}
\end{align}
where \( \mathbf{r}_t \in \mathbb{R}^{N} \) is the reservoir state vector with \( N \) nodes, 
\( \mathbf{x}_t \in \mathbb{R}^{M} \) is the input vector with \( M \) input dimensions, 
\( \mathbf{W}_{\text{res}} \in \mathbb{R}^{N \times N} \) is the fixed internal reservoir weight matrix, 
\( \mathbf{W}_{\text{in}} \in \mathbb{R}^{N \times M} \) is the input weight matrix, 
and \( \mathbf{b} \in \mathbb{R}^{N} \) is a bias vector. 
The nonlinearity is typically chosen as a hyperbolic tangent function.

The reservoir is connected to a linear readout layer, which maps the state vector to the predicted output
\begin{align}
   \bar{\mathbf{y}}_n = \mathbf{W}_{\text{out}} \mathbf{r}_t,
   \label{eq:rc-output}
\end{align}
where \( \mathbf{W}_{\text{out}} \in \mathbb{R}^{Y \times N} \) is the trainable output weight matrix and \( Y \) denotes the number of output dimensions.

For training, the reservoir states are collected over time into a feature matrix
\begin{align}
  \mathbf{R} =
  \begin{bmatrix}
  \mathbf{r}_1^\top \\
  \mathbf{r}_2^\top \\
  \vdots \\
  \mathbf{r}_T^\top
  \end{bmatrix},
  \label{eq:rc-feature-matrix}
\end{align}
where each row corresponds to a time step and each column corresponds to a reservoir node. The output weights are optimized via ridge regression:
\begin{align}
   \mathbf{W}_{\text{out}} =  (\mathbf{R}^\top \mathbf{R} + \lambda \mathbf{I})^{-1}\mathbf{R}^\top \mathbf{Y},
   \label{eq:rc-ridge}
\end{align}
where \( \mathbf{Y} \) contains the desired outputs, \( \lambda \) is a regularization parameter, and \( \mathbf{I} \) is the identity matrix. In this study we use gradient descent for training instead.

\subsection{Attention-Enhanced Reservoir Computing (AERC)}
\label{subsec:aerc-discrete}

Augmenting RC with an attention mechanism replaces the static linear readout by a
state-dependent mapping that adapts to the reservoir state.
Instead of a fixed $\mathbf{W}_{\text{out}}$, we learn a small neural network
$F_{\text{att}}$ with parameters $\mathbf{W}_{\text{net}}$ that produces, at each time~$t$,
a matrix of attention weights $\mathbf{W}_{\text{att},t}$ conditioned on the current
reservoir state $\mathbf{r}_t$:
\begin{align}
  \mathbf{W}_{\text{att},t} &= F_{\text{att}}\!\left(\mathbf{W}_{\text{net}},\, \mathbf{r}_t\right),
  \label{eq:aerc-att-weights}
\\
  \bar{\mathbf{y}}_n &= \mathbf{W}_{\text{att},t}\, \mathbf{r}_t .
  \label{eq:aerc-output}
\end{align}
Here $\mathbf{r}_t \in \mathbb{R}^{N}$, $\mathbf{W}_{\text{att},t} \in \mathbb{R}^{Y\times N}$, and
$\bar{\mathbf{y}}_t \in \mathbb{R}^{Y}$. In this work $F_{\text{att}}$ is a one-hidden-layer MLP
with a ReLU activation; other lightweight parametrizations are possible.
This state-conditional reweighting lets the model emphasize task-relevant coordinates of
$\mathbf{r}_t$ as the underlying regime evolves, improving short-horizon accuracy and long-term statistical fidelity without retraining \cite{vaswani2017attention,bai2018empirical, koester2024}.

Given targets $\mathbf{y}_n$, we minimize a prediction loss over a window of $T_{\text{train}}$
steps (e.g.\ normalized RMSE or MSE):
\begin{align}
  L_{\text{pred}}
   \;=\;
   \frac{1}{N} \sum_{t} \ell\!\left(\bar{\mathbf{y}}_n, \mathbf{y}_n\right),
   \qquad
   \bar{\mathbf{y}}_t = \mathbf{W}_{\text{att},t} \mathbf{r}_t ,
   \label{eq:aerc-loss}
\end{align}
and update $\mathbf{W}_{\text{net}}$ by gradient descent:
\begin{align}
  \mathbf{W}_{\text{net}}^{(s+1)}
  \;=\;
  \mathbf{W}_{\text{net}}^{(s)} \;-\; \gamma \,\nabla_{\mathbf{W}_{\text{net}}} L_{\text{pred}} ,
  \label{eq:aerc-gd}
\end{align}
with epoch index $s$ and learning rate $\gamma$.

\vspace{0.6em}
\subsection{Continuous–Based Reservoir Computing}
\label{subsec:pde-rc}

We now move from node-based (discrete) reservoirs to a spatiotemporal reservoir.
The following formalism can be applied to any continuous reservoir. As an example we will describe a reservoir based on a partial differential equation (PDE). Let $u(\mathbf{x},t)$ denote the reservoir field on a spatial
domain $\Omega \subset \mathbb{R}^d$ ($d\in\{1,2,3,\dots\}$). Its evolution is governed by
\begin{align}
  \partial_t u(\mathbf{x},t)
  \;=\; \mathcal{L}[u](\mathbf{x},t)
        \;+\; \mathcal{I}\!\left(\mathbf{x}; \mathbf{x}_t\right),
  \label{eq:pde-core}
\end{align}
where $\mathcal{L}$ encodes intrinsic transport/reaction dynamics (e.g.\ diffusion
$\nu\Delta u$, advection $\mathbf{c}\!\cdot\!\nabla u$, local reaction $g(u)$), and
$\mathcal{I}$ injects the discrete-time driver $\mathbf{x}_t$ (held constant on $[t,t+\Delta t)$)
through spatial actuators.

To obtain a finite-dimensional state analogous to $\mathbf{r}_t$, we define measurement functionals $\{\phi_t^{(i)}\}_{i=1}^N$, and collect
\begin{align}
  r_t^{(i)} \;=\; \int_{\Omega} \phi_t^{(i)}(\mathbf{x})\, u(\mathbf{x},t)\, d\mathbf{x},
  \qquad
  \mathbf{r}_t = \big(r_t^{(1)},\ldots,r_t^{(N)}\big)^\top .
  \label{eq:pde-meas}
\end{align}
These $\{\phi_t^{(i)}\}_{i=1}^N$ model different forms of measurement procedures that extract a state from the underlying reservoir.
Using the extracted states $r_t^{(i)}$ we can now follow the same logic as described in Sections \ref{subsec:classic-rc} and \ref{subsec:aerc-discrete} to approximate a given target $\mathbf{y}_t$.

\begin{table*}[t]
\centering
\caption{Comparison of classical Reservoir Computing (RC), Attention--Enhanced RC (AERC), and Adaptive-Sensing AERC (ASAERC).}
\label{tab:rc-comparison}
\renewcommand{\arraystretch}{1.15}
\setlength{\tabcolsep}{8pt}
\begin{tabular}{p{3.0cm}p{4.2cm}p{4.2cm}p{4.2cm}}
\toprule
\textbf{Aspect} & \textbf{Classical RC} & \textbf{AERC} & \textbf{ASAERC} \\
\midrule
\textbf{Sensing / readout} &
Static linear map $\bar{\mathbf{y}}_t = \mathbf{W}_{\text{out}}\mathbf{r}_t$. &
State--conditional weights $\mathbf{W}_{\text{att},t} = F_{\text{att}}(\mathbf{r}_t)$ (small MLP) used in prediction, $\bar{\mathbf{y}}_t = \mathbf{W}_{\text{att},t}\mathbf{r}_t$. &
Dual mechanism:
(i) adaptive kernels $\{\phi^{(i)}_{t+T}\}$ select \emph{where to measure};
(ii) attention weights $\mathbf{W}_{\text{att},t+T}$ decide \emph{how to combine}. \\
\textbf{Trainable parameters} &
Only $\mathbf{W}_{\text{out}}$ trained by ridge regression or gradient descent. &
Parameters of $F_{\text{att}}$ trained by gradient descent. &
Parameters of $F_{\theta}$ and $F_{\text{att}}$ (often shared backbone) trained by gradient descent. \\
\bottomrule
\end{tabular}
\end{table*}

\subsection{Adaptive-Sensing Attention-Enhanced Reservoir Computing (ASAERC)}
\label{subsec:caerc-def}

This section explains our new approach in detail.
Attention can play a dual role: it learns both
\textbf{(i) where to sense the field} and \textbf{(ii) how to combine the sensed features into an
output.} To formalize this, we distinguish two types of measurements:

\paragraph*{1. Fixed measurements (inputs to attention).}
At each time step we evaluate a set of fixed measurement functionals
$\{\psi_t^{(j)}(\mathbf{x})\}_{j=1}^{N_\text{fix}}$ on the field $u(\mathbf{x},t)$,
producing a baseline feature vector
\begin{align}
  \tilde{r}_t^{(j)} = \int_{\Omega} \psi_t^{(j)}(\mathbf{x})\, u(\mathbf{x},t)\, d\mathbf{x},
  \qquad
  \tilde{\mathbf{r}}_t = \big(\tilde{r}_t^{(1)},\ldots,\tilde{r}_t^{(N_\text{fix})}\big)^\top .
\end{align}
These fixed features serve as the input state to the attention network.

\paragraph*{2. Adaptive measurements (learned sensors).}
Conditioned on $\tilde{\mathbf{r}}_t$, the network predicts parameters
$\{\theta_{t}^{(i)}\}_{i=1}^{N}$ that define a set of dynamic spatial (or temporal) kernels
$\phi_{t+T}^{(i)}(\mathbf{x})$:
\begin{align}
  \{\theta_t^{(i)}\}_{i=1}^{N}
    &= F_{\theta}\!\left(\mathbf{W}_{\text{net}},\, \tilde{\mathbf{r}}_t\right),
    \qquad
    \phi_{t+T}^{(i)}(\mathbf{x}) = \kappa\!\big(\mathbf{x};\, \theta_t^{(i)}\big),
  \label{eq:caerc-kernel}
\end{align}
where $\kappa$ is a differentiable kernel family (e.g.\ Gaussian bumps, FEM shape functions).
These kernels act as sensors whose locations vary with time.
Their evaluations yield the adaptive reservoir state as follows.
\begin{align}
  r_{t+T}^{(i)} = \int_{\Omega} \phi_{t+T}^{(i)}(\mathbf{x})\, u(\mathbf{x},t+T)\, d\mathbf{x}, \\
  \qquad
  \mathbf{r}_{t+T} = \big(r_{t+T}^{(1)},\ldots,r_{t+T}^{(N)}\big)^\top
\end{align}
These states are analogous to the reservoir state $\mathbf{r}_t$ in classical RC or AERC case.

\paragraph*{3. Adaptive weighting (readout).}\mbox{}\\
The same network then produces a matrix of attention weights that linearly combines these
adaptive measurements:
\begin{align}
  \mathbf{W}_{\text{att},t+T}
    = F_{\text{att}}\!\left(\mathbf{W}_{\text{net}},\, \tilde{\mathbf{r}}_t\right),
    \qquad \\
  \bar{\mathbf{y}}_n = \mathbf{W}_{\text{att},t+T} \mathbf{r}_{t+T} .
  \label{eq:caerc-weights}
\end{align}
Together, $F_{\theta}$ and $F_{\text{att}}$ form a trainable interface that first decides where to look (through $\theta_t^{(i)}$) and then how to combine what it measures.
(through $\mathbf{W}_{\text{att},t+T}$).
For simplicity we make both networks share their weights except for a last separate output head, one for the adaptive measurement locations and the other for the weighting.

If $\phi_{t+T}^{(i)}(\mathbf{x})$=$\{\psi_t^{(j)}(\mathbf{x})\}_{j=1}^{N_\text{fix}}$ and are held fixed in space, the model reduces to AERC
(Sec.~\ref{subsec:aerc-discrete}). 
Replacing the PDE dynamics with a node-based update further recovers the classical AERC for a discrete setting.

\subsection{Flowchart of Continuous Attention-Enhanced Reservoir Computing}
\label{subsec:caerc-flow}

The standardized input $\mathbf{x}_n$ acts as a forcing term for a fixed continuous reservoir, evolving the spatiotemporal field $u(\mathbf{x},t)$.
At each time step, the field is first sampled through a set of fixed kernels $\{\psi_t^{(j)}(\mathbf{x})\}_{j=1}^{N_\text{fix}}$ to form the feature vector
$\tilde{\mathbf{r}}_t = \big(\tilde{r}_t^{(1)},\ldots,\tilde{r}_t^{(N_\text{fix})}\big)^\top$.
A trainable attention module maps these features to two outputs:
(i) kernel parameters $\{\theta_t^{(i)}\}$ that define adaptive measurement kernels and specify \emph{where} the field is sampled,
and (ii) a state-dependent attention weight matrix $\mathbf{W}_{\mathrm{att},t}$ that specifies \emph{how} sampled values are combined.
The final prediction is obtained as $\bar{\mathbf{y}}_n = \mathbf{W}_{\mathrm{att},t+T}\mathbf{r}_{t+T}$.
Crucially, gradients update only the attention network; no gradients are backpropagated through the reservoir.

\subsection{Training of Continuous Attention--Enhanced Reservoir Computer}
\label{seq:loss}

We train ASAERC for one-step-ahead prediction of the system state.
The predicted vector $\bar{\mathbf{y}}_{t}$ is compared with the ground truth $\mathbf{y}_{t}$ using the mean-squared error (MSE):
\begin{align}
  L_{\mathrm{MSE}} = \frac{1}{N} \sum_{n=1}^{N}
  \big\| \bar{\mathbf{y}}_{n} - \mathbf{y}_{n} \big\|^2 ,
  \label{eq:caerc-mse}
\end{align}
where $N$ is the number of training samples.

Gradients are backpropagated through the attention networks, updating the shared parameters $\mathbf{W}_{\text{net}}$:
\begin{align}
  \mathbf{W}_{\text{net}}^{(s+1)}
  = \mathbf{W}_{\text{net}}^{(s)} - \gamma \,
    \nabla_{\mathbf{W}_{\text{net}}} L_{\mathrm{MSE}} ,
  \label{eq:caerc-gd}
\end{align}
where $s$ is the training epoch and $\gamma$ the learning rate.

For all experiments we use the Adam optimizer with an initial learning rate of $0.002$, an exponential weight decay factor of $\gamma = 0.99$, a batch size of $1024$, and a maximum of $500$ epochs.

\begin{table*}[t]
\centering
\caption{Sampling parameters for the eight chaotic systems used in this study. Total time and step size are chosen to ensure convergence to the attractor and to provide sufficient diversity in time scales. The Lyapunov exponents are representative values from the literature.}
\label{tab:sampling_steps}

\setlength{\tabcolsep}{4pt}   % tighter columns
\renewcommand{\arraystretch}{1.1}

\begin{adjustbox}{max width=\textwidth}
\begin{tabular}{l c c c c c}
\toprule
\textbf{System} &
\textbf{Total} &
\textbf{Sample} &
\textbf{Step} &
\textbf{Samples /} &
\textbf{Lyapunov} \\
&
\textbf{Time} &
\textbf{Points} &
\textbf{Size} &
\textbf{Lyap.\ Time} &
\textbf{Exponent} \\
\midrule
Lorenz            & 375  & 7500 & 0.05 & 22.0  & 0.905 \\
Rössler Attractor & 2000 & 7500 & 0.27 & 52.0  & 0.071 \\
Van der Pol       & 1500 & 7500 & 0.20 & 40    & 0.025 \\
Duffing           & 825  & 7500 & 0.11 & 53.9  & 0.177 \\
Double Pendulum   & 2000 & 7500 & 0.27 & 60    & 0.122 \\
Logistic Map      & 7500 & 7500 & 1    & 14.4  & 0.693 \\
Hénon Map         & 7500 & 7500 & 1    & 9.5   & 0.419 \\
Mackey--Glass     & 7500 & 7500 & 1    & 166.7 & 0.006 \\
\bottomrule
\end{tabular}
\end{adjustbox}
\end{table*}

\subsection{Spatiotemporal reservoir based on 2-D diffusion}
\label{sec:pde-example}

To instantiate the ASAERC framework, we use a two-dimensional diffusion equation as the continuous reservoir. The field $u(\mathbf{x},t)$ evolves on a rectangular domain $\Omega=[0,L_x]\times[0,L_y]$ according to
\begin{align}
  \partial_t u(\mathbf{x},t) = \nu \bigl(\partial_{xx} u + \partial_{yy} u\bigr)
  + f(x,y;\mathbf{x}_t),
\end{align}
where $\nu>0$ is the diffusion coefficient and $f(x,y;\mathbf{x}_t)$ injects the discrete-time input through smooth Gaussian sources. We impose homogeneous Dirichlet boundaries, $u=0$ on $\partial\Omega$, so that the field vanishes at the domain edges. The PDE is discretized on a uniform $n_x \times n_y$ grid and advanced in time by an explicit Euler step with
\begin{align}
  u_{i,j}^{n+1} = u_{i,j}^n
  + \Delta t\bigl[\nu\,\Delta u_{i,j}^n + f(x_i,y_j;\mathbf{x}_t)\bigr],
\end{align}
where $\Delta u_{i,j}^n$ is the second-order finite-difference Laplacian. The time step $\Delta t$ satisfies the Courant–Friedrichs–Lewy stability criterion.

\medskip
\noindent\textbf{Fixed input measurements.}
At every time step, we evaluate $u(\mathbf{x},t)$ at a set of $N_\text{fix}$ fixed measurement points
$\{(x^{(k)},y^{(k)})\}_{k=1}^{N_\text{fix}}$ scattered across the domain. If a point does not coincide with a grid node, its value is obtained by bilinear interpolation from the four nearest grid points:
\begin{align}
  u(x^{(k)},y^{(k)}) \approx
  w_{00}\,u_{i,j} + w_{10}\,u_{i+1,j} \nonumber\\
  + w_{01}\,u_{i,j+1} + w_{11}\,u_{i+1,j+1},
  \label{eq:interpolation}
\end{align}
where $w_{00}+w_{10}+w_{01}+w_{11}=1$. These interpolated values form the reservoir state vector
\begin{align}
  \tilde{\mathbf{r}}_t = \bigl(u(x_t^{(1)},y_t^{(1)}), \dots, u(x_t^{(N_\text{fix})},y_t^{(N_\text{fix})})\bigr)^\top,
\end{align}
which is passed to the attention network $F(\tilde{\mathbf{r}}_t)$.

\medskip
\noindent\textbf{Adaptive output measurements.}
Given $\tilde{\mathbf{r}}_t$, the network outputs a new set of measurement locations
$\{(x_{t+T}^{(i)},y_{t+T}^{(i)})\}_{i=1}^{N}$ that may vary at each time step, along with the attention weights
$\mathbf{W}_{\mathrm{att},t+T}$ used to combine them into the predicted output. The field is again sampled at these locations (using the same interpolation rule), producing
\begin{align}
  \mathbf{r}_{t}
  = \bigl(u(x_{t+T}^{(1)},y_{t+T}^{(1)}),\dots,u(x_{t+T}^{(N)},y_{t+T}^{(N)})\bigr)^\top,
\end{align}
which is linearly combined as $\bar{\mathbf{y}}_t = \mathbf{W}_{\mathrm{att},{t+T}}\mathbf{r}_{{t+T}}$.  
In this way, the fixed measurements provide a stable and consistent input representation, while the adaptive measurements allow the network to dynamically focus on informative regions of the PDE field as time evolves.
The time shift $T$ emulates the information propagation time of the measuring procedures from the time of measuring the reservoir, over processing the information in the attention module to the time of the measuring device to react.

\subsection{Benchmark systems and sampling parameters}
\label{subsec:benchmarks}

We evaluate ASAERC on the task of simultaneously predicting a diverse set of chaotic attractors with a single shared model.
Extending our earlier discrete AERC study \cite{Koester2025}, we increase both the number and heterogeneity of benchmark systems by combining continuous-time ordinary differential equations (ODEs), discrete maps, and a delay differential equation (DDE).
This diversity tests whether continuous spatial attention can adapt across different time scales and dynamical regimes.

Specifically, the benchmark dataset includes:
\begin{itemize}
  \item \textbf{Continuous-time chaotic systems:} the Lorenz system \cite{lorenz1963deterministic}, 
        Rössler system \cite{rossler}, Van der Pol oscillator, Duffing oscillator \cite{duffing}, 
        and the double pendulum. These are integrated using an adaptive Runge--Kutta method.
  \item \textbf{Discrete maps:} the Logistic map and the Hénon map \cite{henon}.
  \item \textbf{Delay differential system:} the Mackey--Glass equation \cite{mackey}, 
        integrated using a discretized delay buffer with Euler stepping.
\end{itemize}

Each system exhibits a distinct attractor geometry. Lorenz System and Rössler Sytem are three-dimensional chaotic flows with different folding mechanisms; Duffing Oscillator and Van der Pol Oscillator exhibit nonlinear oscillations; the double pendulum produces chaotic mechanics under conservative dynamics; Logistic and Hénon maps represent discrete-time chaos; and Mackey-Glass equation introduces long-range memory via delay.
Together, these benchmarks provide a stringent test of representational capacity and adaptivity.

The sampling parameters used in our experiments are summarized in Table~\ref{tab:sampling_steps}.
We choose total integration times and step sizes to ensure convergence to the attractor and to cover diverse time scales.
Each trajectory is resampled to 7500 points for uniformity across systems, which results in a specific step size shown in Table~\ref{tab:sampling_steps}, which is different from the step size used for simulation.
After gathering all eight different attractors data we concatenate them into one long time series and split them into input and target, where target is a the time series shifted by one, i.e. a one step ahead prediction.

\clearpage
\appendix
\section*{Appendix}

\section{Comparison with delay-embedding MLP and LSTM}
\label{subsec:delay-embedding}

We compare against classical RC, static attention variants, and data-driven baselines based on delay-embedding multilayer perceptrons (MLPs) and LSTMs \cite{Packard1980,Takens1981}. 
In the delay-embedding approach, the input at time $t$ is augmented with $k$ delayed copies $(\mathbf{x}_{t-1}, \ldots, \mathbf{x}_{t-k})$, forming a delay vector that is passed into a two-layer MLP trained to predict $\mathbf{y}_t$ directly.

\begin{figure}[t]
\centering
\includegraphics[width=0.48\textwidth]{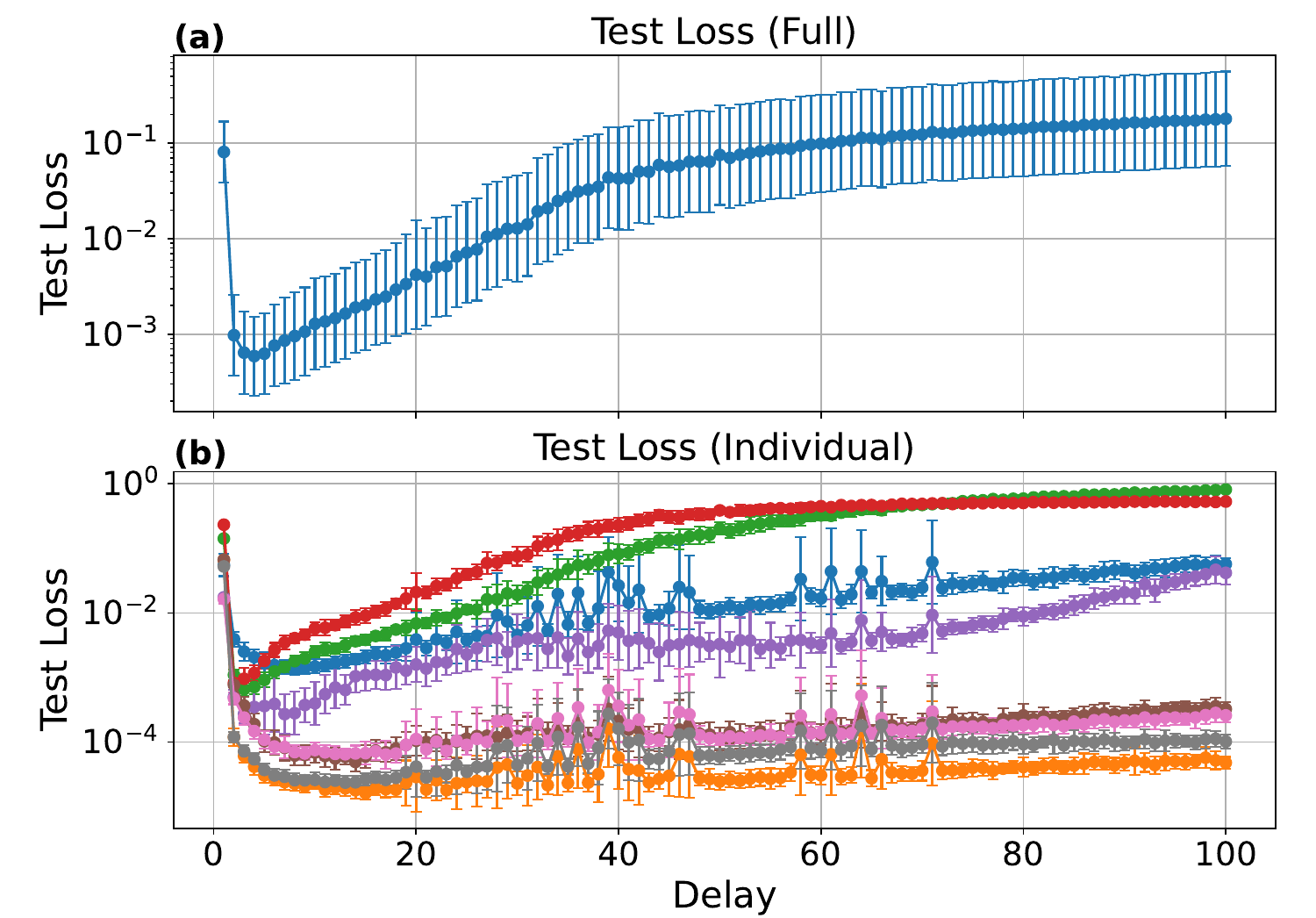}
\caption{\textbf{Delay-embedding MLP baseline.}
Full (all eight systems) and individual normalized MSE versus number of delay steps $k$ and parameter count. Double Pendulum (blue), Duffing Oscillator (orange), Henon Map (green), Logistic Map (red), Lorenz System (purple), Mackey-Glass Equation (brown), Roessler System (pink), Van der Pool Oscillator (grey). 
Performance improves for small $k$ however worsens beyond an optimal value, likely due to increasing input dimensionality and insufficient training or overfitting.}
\label{fig:delay-embedding}
\end{figure}

Figure~\ref{fig:delay-embedding} reports the normalized MSE as a function of delay length $k$, together with the total number of trainable parameters of the resulting network. 
For small $k$, the delay-embedding model yields a moderate reduction in error compared to a single-step input. 
This suggests that a short history helps the network resolve local ambiguities in the dynamics, reaching its lowest error rates at a level comparable to the best ASAERC models. 
However, as $k$ grows beyond an intermediate range, performance deteriorates: the model must process increasingly high-dimensional inputs, which likely require longer training to fit effectively and are prone to overfitting when the number of epochs is fixed. 
Moreover, its performance exhibits substantial instability across runs, reflected by large error bars. 

\begin{figure}[t]
\centering
\includegraphics[width=0.49\textwidth]{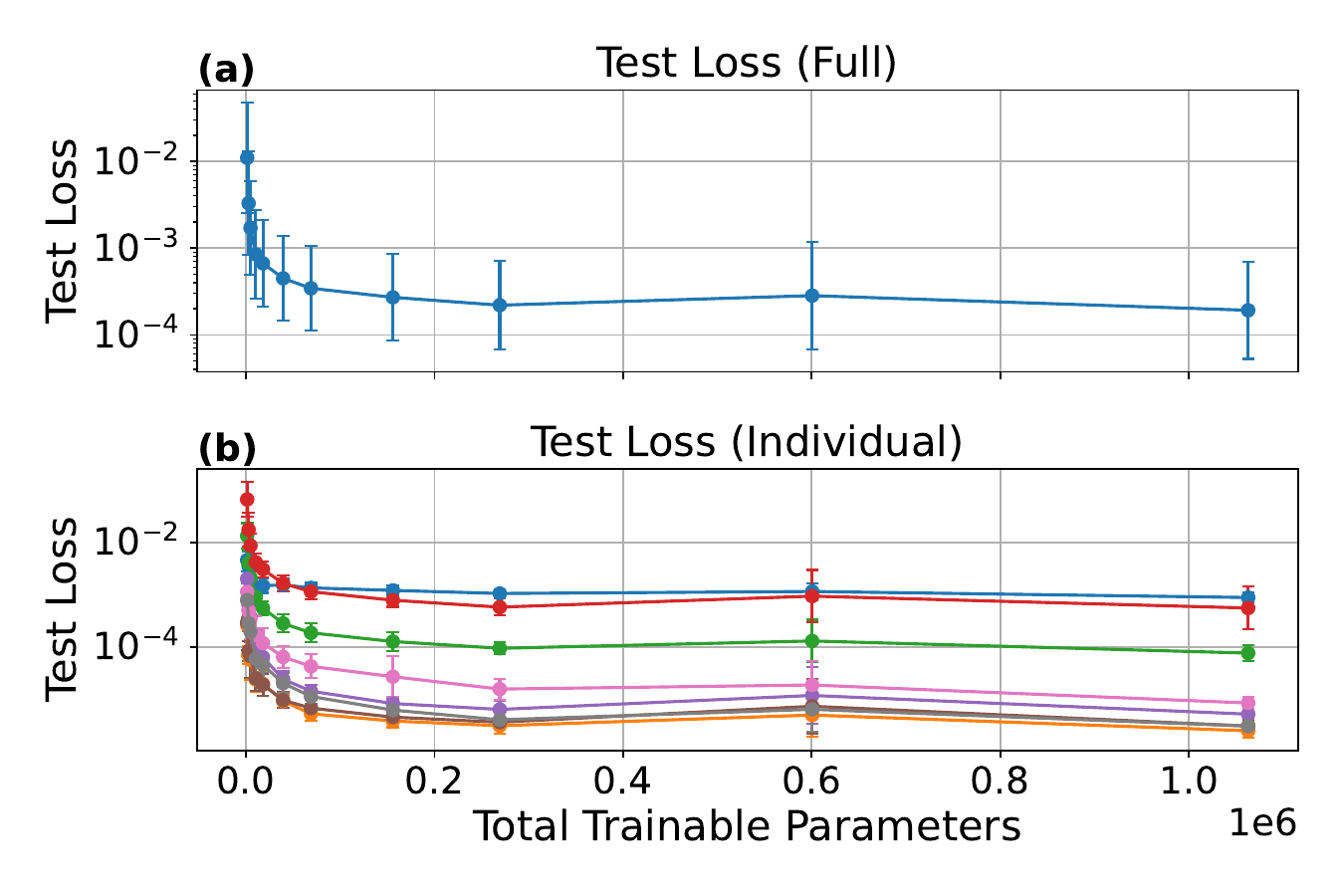}
\caption{\textbf{LSTM baseline.}
Full (all eight systems) and individual normalized MSE versus total parameter count and parameter count over hidden size.
Performance improves monotonically for higher parameter counts, showing a more stable behaviour than the delay-embedding approach. 
The performance is on par with the best ASAERC models while also having roughly the same parameter count. Double Pendulum (blue), Duffing Oscillator (orange), Henon Map (green), Logistic Map (red), Lorenz System (purple), Mackey-Glass Equation (brown), Roessler System (pink), Van der Pool Oscillator (grey).}
\label{fig:lstm_loss}
\end{figure}

In Figure~\ref{fig:lstm_loss}, we show the performance of an LSTM baseline.
The normalized MSE is plotted against the total parameter count.
The results show that performance improves with increasing network size and then reaches a plateau beyond roughly $10^{5}$ parameters.
At this point, the error is slightly better compared to the best ASAERC configurations, and both approaches require a similar number of parameters.
Nevertheless, we point out that ASAERC is a first approach setup and thus has no optimizations done to it all.
Additionally the ASAERC opens up the possibility of using any continuous system as its reservoir where some systems probably yield drastic improvements over the simple diffusion based system used in this paper.
Thus we are surprised that even this simple first approach can get close in its performance to a highly optimized LSTM of similar parameter counts.

\section{Individual Systems MSE}
\label{subsec:ISMSE}

\begin{figure*}[t]
\centering
\includegraphics[width=\textwidth]{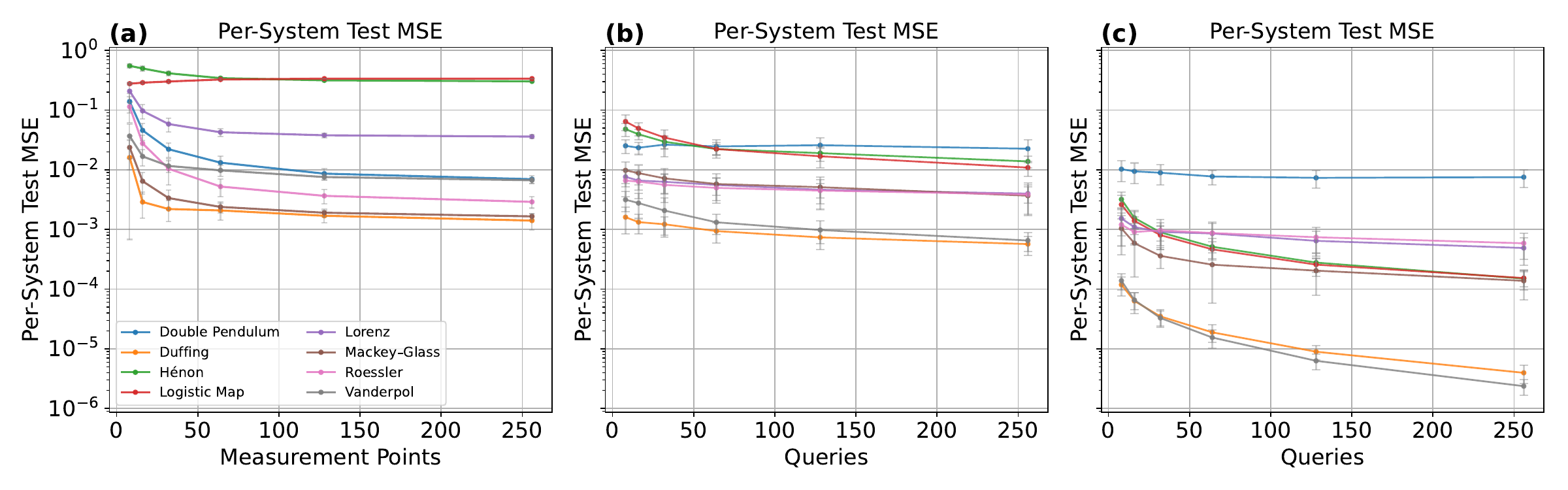}\hfill
\caption{Fig.~\ref{fig:mse-vs-points} MSE for the individual dynamical systems.}
\label{fig:mse-vs-points_app}
\end{figure*}

Figure \ref{fig:mse-vs-points_app} shows the same MSE from Fig. \ref{fig:mse-vs-points} however split into the individual systems of the data set.
From Fig. \ref{fig:mse-vs-points_app} we can see that some systems, like the Duffing oscillator improves dratically over multiple orders of magnitude via the new ASAERC approach.

% ============================================================
\section*{Acknowledgements}
This study was supported in part by JSPS KAKENHI (JP22H05195,
JP24KF0179, JP25H01129) and JST CREST (JPMJCR24R2) in Japan.

% ---------- Bibliography ----------
% For Nature Comms submission you will later paste the .bbl content here.
\bibliographystyle{unsrt}
\bibliography{references}

\end{document}